\def\year{2022}\relax
\documentclass[letterpaper]{article} 
\usepackage{aaai22}  
\usepackage{times}  
\usepackage{helvet}  
\usepackage{courier}  
\usepackage[hyphens]{url}  
\usepackage{graphicx} 
\usepackage{natbib}  
\usepackage{caption} 
\DeclareCaptionStyle{ruled}{labelfont=normalfont,labelsep=colon,strut=off} 
\usepackage{algorithm}
\usepackage{algorithmic}
\usepackage{epsfig}
\usepackage{amsmath}
\usepackage{amssymb}
\usepackage[utf8]{inputenc} 
\usepackage{mathrsfs}
\usepackage{multirow}
\usepackage{enumitem}

\DeclareMathOperator{\E}{\mathbb{E}}
\usepackage{bbm}
\DeclareMathOperator*{\argmax}{arg\,max}

\newcommand{\figref}[1]{Fig.~\ref{#1}}
\newcommand{\tabref}[1]{Table.~\ref{#1}}

\newcommand{\algref}[1]{Algorithm.~\ref{#1}}
\usepackage{color}
\usepackage[table,xcdraw]{xcolor}
\usepackage{colortbl}

\usepackage{lipsum}
\title{Unsupervised Domain Adaptation for Video Semantic Segmentation}

\author{%
  Kwanyong Park\thanks{equal contribution}\quad%
  Inkyu Shin\footnotemark[1]\quad%
  Sanghyun Woo\quad%
  In So Kweon
\\
KAIST, South Korea. \\
{\{pkyong7,dlsrbgg33,shwoo93,iskweon77\}@kaist.ac.kr}
\
}
\affiliations{
    \setcounter{figure}{0}
    \includegraphics[width=0.91\textwidth]{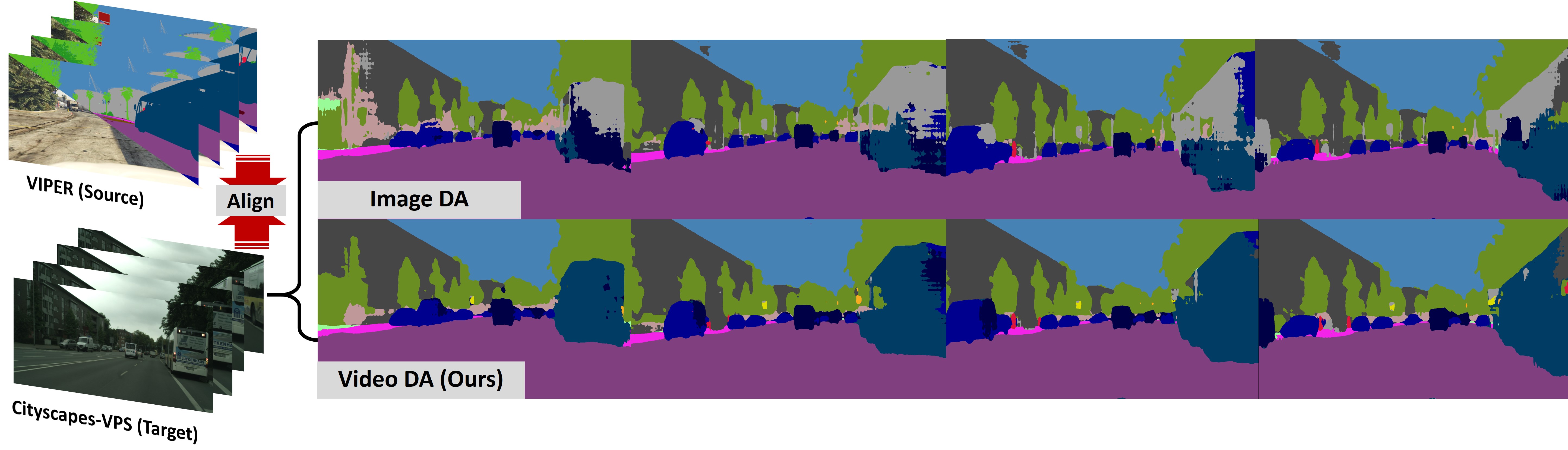}
    \captionof{figure}{\textbf{Adaptation outputs on VIPER to Cityscapes-VPS.} We compare our method with the state-of-the-art Image DA method~\cite{mei2020instance}. It is readily apparent that our predictions are more semantically accurate and temporally consistent.}
}

\begin{document}

\maketitle


\begin{abstract}
Unsupervised Domain Adaptation for semantic segmentation has gained immense popularity since it can transfer knowledge from simulation to real (Sim2Real) by largely cutting out the laborious per pixel labeling efforts at real.
In this work, we present a new video extension of this task, namely \textbf{Unsupervised Domain Adaptation for Video Semantic Segmentation}. 
As it became easy to obtain large-scale video labels through simulation, maximizing Sim2Real knowledge transferability is one of the promising directions for resolving the fundamental data-hungry issue in the video.
To tackle this new problem, we present a novel video-specific two-phase adaptation scheme.
In the first step, we train the model using labeled source videos and align the features from source to target with the guidance of video context.
In the second step, we train the model on target videos using pseudo labels.
To obtain dense and accurate pseudo labels, we exploit the temporal information in the video.
By combining these two steps, we construct a solid video adaptation method.
On the `VIPER to CityscapeVPS' adaptation scenario, we show that our proposals significantly outperform all the previous image-based UDA methods both on image-level (mIoU) and video-level ($VPQ^{s}$) evaluation metrics.
\end{abstract}

\section{Introduction}
A pixel-level understanding of different semantics, object instances, and temporal changes in a video provides primitive information for various vision applications such as autonomous driving, robot control, and augmented reality.
Video semantic segmentation has been regarded as one of the representative proxies for this challenging goal.
While a large number of learning-based approaches under two main research directions of improving accuracy~\cite{nilsson2017semantic, li2018lowlatency} and efficiency~\cite{shelhamer2016clockwork, zhu2017deep} have been presented, the scarcity of the available video labels turns out to be the bottleneck of further advances in this field.

Recently, simulators based on video game engines, which can quickly generate an unlimited amount of high-quality, diverse video labels, emerged as an attractive alternative for laborious human labeling.
However, the model trained on simulated data suffers from the fundamental \textit{domain shift}~\cite{sankaranarayanan2018learning}, and thus the severe performance drop in the real world is inevitable without proper adaptation. To remedy these distribution shifts, the community has put a great effort into developing domain adaptation techniques.  While there has been a flurry of advances, most of the previous studies overlook temporal context that limits adaptation still to an \textit{image level} and thus cannot scale to video model adaptation.

In this work, we present a novel video-specific adaptation method that is comprised of two phases.
The first phase is an adversarial video training (VAT) step that uses video context to train the model and align the features from source to target.
Specifically, a tube matching loss is introduced to explicitly capture temporal changes of shape and boundaries of each segment.
Also, a sequence discriminator is presented to adapt the features better using spatio-temporal information.
The second phase is a video self-training (VST) step that focuses on the target data learning with the generated pseudo labels.
To obtain more dense and accurate pseudo labels, we aggregate information from the neighboring frames.
We show that these two phases are complementary and form a compelling video adaptation method, which sets strong baseline scores on challenging VIPER~\cite{richter2017playing} to CityscapeVPS~\cite{kim2020vps} adaptation scenario.
We summarize our contributions as follows:
\begin{enumerate}
\vspace{-1mm}
\setlength\itemsep{0.3em}
    \item We firstly define and explore unsupervised domain adaptation for video semantic segmentation.
    \item We design a novel two-phase video domain adaptation scheme. The two main components are Video Adversarial Training (VAT) and Video Self Training (VST).
          We show that both are essential and complementary in constructing a compelling video adaptation method.
    \item We validate our proposal on the Viper to Cityscape-VPS scenario. We show that our method significantly outperforms all the previous image-based UDA approaches both on image-specific (mIoU) and video-specific (VPQ~\cite{kim2020vps}) evaluation metrics. Our results clearly show the importance of developing video-specific adaptation techniques.
\end{enumerate}

\section{Related Works}
\subsection{Domain Adaptive Semantic Segmentation}
A number of domain adaptation techniques have been studied in various vision tasks.
Among them, domain adaptive semantic segmentation, which requires the challenging adaptation of structured output at a pixel level.
There are two dominant paradigms in this field: \textit{Adversarial learning-based}~\cite{tsai2018learning, vu2019advent, wang2020differential} and \textit{Self-training-based}~\cite{zou2018domain, zou2019confidence}.
Some works investigated combining these two methods to reduce the domain gap further~\cite{li2019bidirectional, pan2020unsupervised, shin2020two, mei2020instance}.
However, we note these methods are still limited at \textit{image-level} adaptation, which cannot scale to video-level adaptation as it overlooks important temporal information.
Recently, \cite{chen2020generative} attempted to utilize information across the frames for binary segmentation of medical images.
However, its framework neither aims for video segmentation nor employs explicit temporal feature aggregations, which we empirically show that it is inferior to our framework on complex urban scene video segmentation\footnote{We provide detailed analysis in the supplementary material.}.
In this work, we carefully designed a two-phase video adaptation framework that \textit{explicitly} aggregates and utilizes temporal information, which has been rarely studied in earlier works.

\subsection{Video Semantic Segmentation}
A dense understanding of video is one of the long-standing fundamental vision problems that have various applications.
As a representative proxy, video semantic segmentation has been actively studied to improve accuracy and efficiency.
However, most current video segmentation benchmarks provide annotations for only a single frame per video clip~\cite{Cordts2016Cityscapes, yu2020bdd100k}, mainly due to laborious dense labeling efforts.
We instead explore video simulated learning, which allows us to avoid expensive manual labeling, and propose a strong video adaptation method.

\section{Preliminary}
For better understanding of our proposals, we provide preliminaries based on the state-of-the-art image domain adaptive segmentation framework, IAST~\cite{mei2020instance}.

\subsection{Problem Setting in image UDA}

We assume that both the data and labels, $(x_{s},y_{s}) \in (\textbf{X}_{\textbf{S}},\textbf{Y}_{\textbf{S}})$, are available in source domain.
In contrast, the data, $x_{t} \in \textbf{X}_{\textbf{T}}$, is only available in target domain.

\subsection{Image Adversarial training(IAT)}
Adversarial training in image DA can be conducted via minimizing source-specific segmentation loss and generative adversarial loss. 
We denote G, $P_{s}=G(x_{s})$ and $P_{t}=G(x_{t})$ as semantic segmentation network, predicted source, and target segmentation map, respectively. 
Source-specific segmentation loss is defined as $L_{seg}(P_{s},y_{s})$, where $L_{seg}$ is cross-entropy.

To align the distribution between source and target image, adversarial learning~\cite{tsai2018learning} is applied.
\begin{equation}
    \begin{split}
    &\mathcal{L}_{Adv,I}(G,D_{I})=
    \E_{\mathbf{x}_{s} \sim X_{S}}\left [ \log D_{I}(P_{s}) \right ] \\
    &+ \E_{\mathbf{x}_{t} \sim X_{T}}\left [ \log(1-D_{I}(P_{t})) \right ]
    \end{split}
    \label{eqn:img_Gan_Loss}
\end{equation}
A image discriminator $D_{I}$ tries to discriminate between target segmentation map $P_{t}$ and source segmentation map $P_{s}$, while the segmentation network G attempts to fool the discriminator.
Therefore, full objective function in IAT can be summarized as below:
\begin{equation}
\begin{split}
L_{IAT}(G,D_{I})= L_{seg}(P_{s},y_{s}) + \lambda_{Adv} \mathcal{L}_{Adv,I}(G,D_{I})
\end{split}
\label{eqn:total_IAT}
\end{equation}

\subsection{Image Self training(IST)}
Self-training iteratively generates pseudo labels and re-trains the model. 
In typical, the pseudo label $\hat{y}_{t}$ can be formulated as
\begin{equation}
  \hat{y}^{(k)}_{t}= \argmax_{k \in K}\mathbbm{1}_{P^{(k)}_{t} >  \theta^{(k)}} P^{(k)}_{t}
  \label{eq:case1}
\end{equation}
where $\mathbbm{1}$ is a function which returns the input if the condition is true or ignored otherwise.
$\hat{y}^{(k)}_{t}$ and $P^{(k)}_{t}$ are pseudo label and prediction for class k, respectively. 
When the confidence is above a certain threshold $\theta^{(k)}$, the predicted class is selected as a pseudo label. 

Defining proper threshold $\theta^{(k)}$ is crucial for cutting out the noisy pseudo labels.
In practice, IAST~\cite{mei2020instance} adjusts it for each images to diversify the pseudo labels.
The threshold of current image $\theta_{t}^{(k)}$, is decided as follows:
\begin{equation}
\begin{split}
\theta_{t}^{(k)} = \beta\theta_{t-1}^{(k)} + (1-\beta)\Psi(x_{t}, \theta_{t-1}^{(k)}), \\
\Psi(x_{t}, \theta_{t-1}^{(k)}) = \mathbbm{P}_{t}^{(k)}[\alpha{\theta_{t-1}^{(k)}}^{\gamma}|\mathbbm{P}_{t}^{(k)}|]
\end{split}
\label{eqn:adapt_select}
\end{equation}
$\theta_{t}^{(k)}$ is a moving averaged threshold with a momentum factor $\beta$.
$\Psi(x_{t}, \theta_{t-1}^{(k)})$ represents the threshold for acquring the current instance.
$\alpha$ controls the proportion of pseudo labels, and $\gamma$ is a weight decay parameter to reduce pseudo labels with low confidence.


\begin{figure*}[t]
    \centering 
    \includegraphics[width=0.95\textwidth]{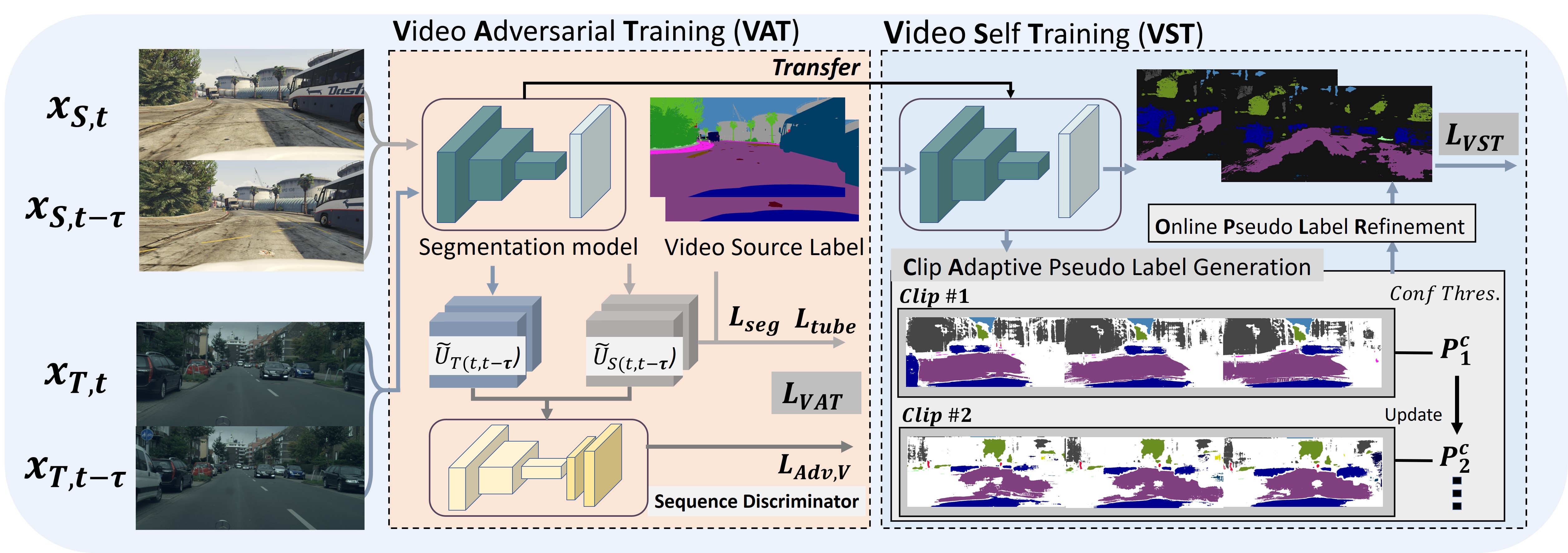}
    \caption{\textbf{The overview of the proposed unsupervised domain adaptation for video semantic segmentation(Video DA).} Our Video DA consists of two phase video specific domain adaptation training: Video Adversarial Training(VAT) and Video Self Training(VST). In first phase, VAT, we stack two neighbor outputs from two domains(${U}_{S(t,t-\tau)}$ for source and ${U}_{T(t,t-\tau)}$ for target) and utilize sequence discriminator to align them. In second phase, VST, we design a aggregation based clip adaptive pseudo label generation strategy.}
    \label{fig:main_figure}
    \vspace{-3mm}
\end{figure*}

Given pseudo labels on the target domain, the image self-training loss can be defined as:
\begin{equation}
\begin{split}
L_{IST}(G)= \lambda_{seg,s}L_{seg}(P_{s},y_{s}) \\
+\lambda_{seg,t}L_{seg}(P_{t},\hat{y}_{t}) + \lambda_{reg}L_{reg}(P_{t},\hat{y}_{t})
\end{split}
\label{eqn:total_IST}
\end{equation}
The third term represents additional regularization for self-training~\cite{zou2019confidence, mei2020instance}. 
For more details, please refer to the original papers.

\section{Methods}

\subsection{Problem Setting in Video UDA}
Video domain adaptation aims to transfer space-time knowledge from labeled source domain S to the unlabeled target domain T. 
Different from the single frame setting in image UDA,
source set consists of temporally ordered frame sequences, $x_{S} := \{x_{S,1},x_{S,2},...,x_{S,t},...\}$ $(x_{S} \in \textbf{X}_{\textbf{S}})$, and corresponding label sequences, $y_{S} := \{y_{S,1},y_{S,2},...,y_{S,t},...\}$ $(y_{S} \in \textbf{Y}_{\textbf{S}})$. 
In the target domain, we have data without labels. 
We denote target frame sequences as $x_{T} := \{x_{T,1},x_{T,2},...,x_{T,t},...\}$ $(x_{T} \in \textbf{X}_{\textbf{T}})$.


\subsection{Overview of Proposed Method}
Unlike static images, videos have rich temporal and motion information. 
To train a model with such information, we propose a novel two-phase approach that is consisted of video adversarial training(VAT) and video self-training(VST). 
In the VAT step, the model learns source domain knowledge using video labels, and adapts its features from source to target. 
In the VST step, the model directly learns target domain knowledge based on the generated pseudo labels.
We note that \textit{video context} is explicitly utilized in both steps.
\figref{fig:main_figure} shows the overview of the proposed method. 
We detail each step in the following sections.

\subsection{Video Adversarial Training(VAT)}
In this step, we aim to learn source domain knowledge using video labels and adapt the features from source to target via adversarial loss.
Specifically, given input source frames $x_{S,t}$, ${x}_{S,t-\tau}$ from time $t$ and $t-\tau$, the image semantic segmentation model $G$ predicts ``soft-segmentation map" $P_{S,t}, P_{S,t-\tau}$. A conventional frame-level semantic segmentation is learned from applying the standard cross-entropy loss to each frames.

We further leverage source labels and designed a new video level supervision signal, tube matching loss, that can capture the finer temporal changes of each segment.
Also, to facilitate the adversarial learning with the additional temporal context, we present sequence discriminator. We detail each in the following.


\subsubsection{Tube Matching Loss}
Training a model with each frame independently apparently misses rich temporal information.
Here, we propose a novel tube matching loss that compares the tube prediction of each class with the groundtruth tube, which aims to capture and correct the boundary and shape changes of every segment in the video.
We consider two-length tube in this paper due to the GPU memory limit.
In particular, we denote $\widetilde{U}_{S(t,t-\tau)} = [P_{S,t}, P_{S,t-\tau}]$ as the sequence of source prediction that is a concatenation of two soft segmentation maps in temporal dimension. $\widetilde{U}_{S(t,t-\tau)}^{(k)} =[P_{S,t}^{(k)}, P_{S,t-\tau}^{(k)}]$ is a segment tube of class k.
Groundtruth tube ${U}_{S(t,t-\tau)}^{(k)}$ is defined similarly.
The tube matching loss is instantiated with the dice coefficient~\cite{milletari2016vnet} between the prediction $\widetilde{U}_{S(t,t-\tau)}$ and groundtruth ${U}_{S(t,t-\tau)}$.
\begin{equation}
    \begin{split}
    \mathcal{L}_{tube}\mathrm(\widetilde{U}_{S(t,t-\tau)}, {U}_{S(t,t-\tau)}) = \\
    \sum_{k=1}^K [1 - Dice(\widetilde{U}_{S(t,t-\tau)}^{(k)}, {U}_{S(t,t-\tau)}^{(k)})]\\
    \end{split}
    \label{eq:tube}
\end{equation}
The dice coefficient is insensitive to number of each class prediction and defined as 
\begin{equation}
    \begin{split}
    Dice(p,q) = \frac{2\sum(p q)}{\sum p^2 + \sum q^2}
    \end{split}
    \label{eq:dice}
\end{equation}
We empirically show that the tube matching loss allows the model to learn more discriminative feature compared to the baseline model trained only with the standard frame-level cross-entropy loss, indicating that incorporating the temporal context in feature learning is crucial.

\subsubsection{Sequence Discriminator}
To alleviate domain discrepancy both in space and time,  we propose to use sequence discriminator. Different from image discriminator~\cite{tsai2018learning,vu2019advent}, sequence discriminator $D_{V}$ takes sequence of soft segmentation maps as an input. The discriminator is trained to discriminate sequence of source and target. At the same time, the segmentation model attempts to fool the discriminator. The video GAN loss is as follows:
\begin{equation}
    \begin{split}
    L_{Adv,V}(G,D_{V})= 
    \E_{\mathbf{x}_{S,t},\mathbf{x}_{S,t-\tau} \sim X_{S}}\left [ \log D_{V}(\widetilde{U}_{S(t,t-\tau)}) \right ] \\
    + \E_{\mathbf{x}_{T,t},\mathbf{x}_{T,t-\tau} \sim X_{T}}\left [ \log(1-D_{V}(\widetilde{U}_{T(t,t-\tau)})) \right ]
    \end{split}
    \label{eqn:Vid_Gan_Loss}
\end{equation}
where $\widetilde{U}_{T(t,t-\tau)}$ is sequence of prediction on target images, $x_{T,t}$ and ${x}_{T,t-\tau}$.
By construction, the semantic segmentation model learns to align sequence distribution between source and target.

\subsubsection{Full objective function for VAT} Total loss function is defined as
\begin{equation}
\begin{split}
L_{VAT}(G,D_{v})= \sum_{l=t,t-\tau}[\mathcal{L}_{seg}(P_{S,l},y_{S,l})] \\
+ \mathcal{L}_{tube}(\widetilde{U}_{S(t,t-\tau)}, {U}_{S(t,t-\tau)}) + \lambda_{V} \mathcal{L}_{Adv,V}(G,D_{V})
\end{split}
\label{eqn:total}
\end{equation}
Along with per-frame cross entropy loss, the tube matching loss enforce model to produce accurate and consistent prediction on source domain, while video gan loss adapts the learned representation to the target domain.

\subsection{Video Self Training(VST)}


Different from previous methods~\cite{zou2018domain, mei2020instance} that treat each image independently, we predict pseudo label at the clip level as a whole which leads to robustly cut-out the noisy predictions. In the training time, we additionally conduct the pseudo label refinement step to further eliminate error in the pseudo label.
To benefit from temporal information both in \textit{pseudo label generation} and \textit{pseudo label refinement}, we aggregate predictions and employ temporal consensus of pseudo label across the different time steps. 


\begin{algorithm}[t]
\caption{Pseudo Label Generation}
\label{alg:pseudo}
\textbf{Input}: Model G, $\bigl[x_{T} = \{x_{T,1},x_{T,2},...,x_{T,t},...\}\bigr]^{N}_{T=1}$ \\
\textbf{Parameter}: proportion $\alpha$, momentum $\beta$ \\
\textbf{Output}: target video pseudo-labels \\
  \begin{algorithmic}[1]
    \STATE \textbf{init $\theta_{0}$=0.9}
    \FOR{T=1 to N}
        \STATE $\widetilde{U}_{T(1:N)}^{A} = \textit{Aggregate}(G(x_{T}))$ -//* Eq.~\eqref{eqn:agg} *//- \\
        \STATE $\widetilde{U}_{index} = \argmax(\widetilde{U}_{T(1:N)}^{A})$ \\
        \STATE $\widetilde{U}_{value} = \max(\widetilde{U}_{T(1:N)}^{A})$
        \FOR{$k$ in $K$}
            \STATE $\widetilde{\mathbbm{U}}_{T}^{(k)}$ = $sort(\widetilde{U}_{value}[\widetilde{U}_{index} = k])$ \\
            \STATE $\theta_{x_{T}}^{(k)}$ = $\Psi(x_{T}, \theta_{T-1}^{(k)})$ -//* Eq.~\eqref{eqn:CAST_select} *//- \\
        \ENDFOR
        \STATE $\boldsymbol{\theta_{T}}$ = $\beta\boldsymbol{\theta_{T-1}} + (1-\beta)\boldsymbol{\theta_{x_{T}}}$  -//* Eq.~\eqref{eqn:CAST_select} *//- \\
        \STATE $\boldsymbol{\hat{y}_{T}}$ = $onehot(\widetilde{U}_{index}[\widetilde{U}_{value} > \boldsymbol{\theta_{T}}])$ \\
    \ENDFOR
    \STATE $\textbf{return {$\bigl[\hat{y}_{T}\bigl]^{N}_{T=1}$}}$
  \end{algorithmic}
\end{algorithm}

\begin{figure}[t]
    \centering 
    \includegraphics[width=0.49\textwidth]{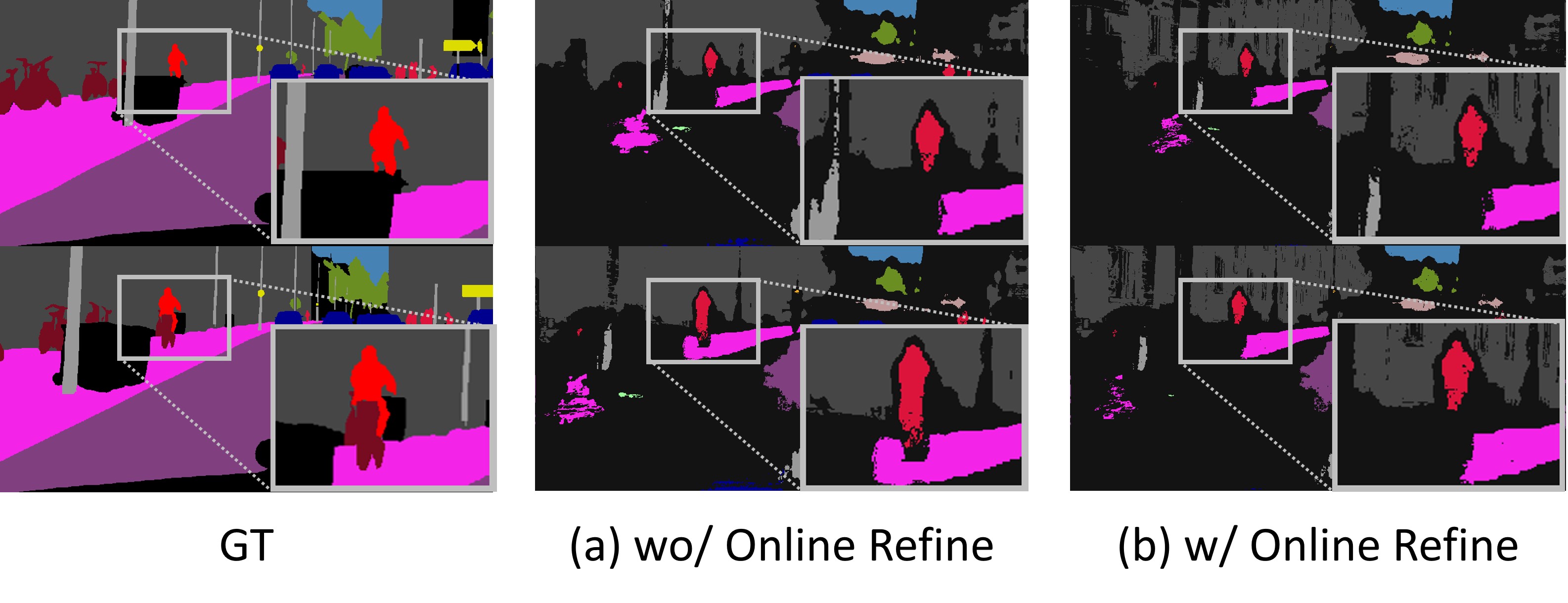}
    \vspace{-6mm}
    \caption{\textbf{Visualization of pseudo-labels w/o and w/ the online refinement.} The proposed online refinement successfully eliminate noise on the pseudo labels by checking temporal consensus.}
    \label{fig:online_qual}
    \vspace{-3mm}
\end{figure}

\subsubsection{Temporally Aggregated Prediction}
In pseudo label generation step, we take two neighbor frames $x_{T,j}$  ($j\in\{i-1,i+1\}$), to aggregate their predictions onto a target frames, $x_{T,i}$.
Given a target frame $x_{T,i}$ and a neighbor frame $x_{T,j}$, pixel-level correspondence is estimated by an optical flow network(e.g. FlowNet2.0~\cite{ilg2016flownet}). 
The predicted soft-segmentation maps on the neighbor frame are warped to the target frame as follows. 
To aggregate diverse prediction from multiple frames, we average the predictions from target and neighbor frame when the warped prediction is not occluded. 
The non-occluded map from $x_{T,j}$ to $x_{T,i}$ is defined by $\mathbb{I}_{j\Rightarrow i} = \{\boldsymbol{1}|O_{j \Rightarrow i} > th \}$ where $th$ is threshold value and $O_{t \Rightarrow t-1}=\exp(-\alpha||x_{T,i}-W_{j \Rightarrow i}({x}_{T,j})||_2)$
denotes the occlusion mask which is calculated from the warping error between target frame $x_{T,i}$ and warped neighbor frame $W_{j \Rightarrow i}({x}_{T,j})$. The aggregated prediction at target frame is computed as,
\begin{equation}
P_{T,i}^{A} = \frac{{P}_{T,i} + \sum\limits_{j} \mathbb{I}_{j\Rightarrow i} W_{j \Rightarrow i}({P}_{T,j})}{1 + \sum\limits_{j} \mathbb{I}_{j\Rightarrow i}}
\label{eqn:agg}
\end{equation}
Thus, sequence of aggregated prediction for a clip $x_{T}$ is expressed as, $\widetilde{U}_{T(1:N)}^{A} = [P_{T,1}^{A},P_{T,2}^{A},...,P_{T,t}^{A},...]$ where N represent total frame number of each clip. We abbreviate above process as \textit{Aggregate}.

\begin{table*}[t]
\centering
\resizebox{\textwidth}{!}{\begin{tabular}{l | c | c c c c c c c c c c c c c c c | c }
\hline 
\hline
\multicolumn{18}{c}{VIPER $\rightarrow$  Cityscapes-VPS [mIoU]} \\
\hline
Method &Arch. &RD &SW & BD & FN & PL &TL & TS &Veg. &TR  &SKY & PR & Car& Truck &Bus &Motor & mIOU(\%) \\
\hline
Source-only & - &44.24 &22.70 &73.69 &6.02 &6.74 &10.47 &19.25 &82.55 &31.13 &80.74 &60.91 &62.37 &2.25 &7.18 &0.0 & 34.02\\
\hline
AdaptSegNet~\cite{tsai2018learning} & AT &86.09 &41.54 &80.59 &16.10 &12.71 &20.00 &23.98 &82.44 &34.05 &80.14 &64.79 &69.23 &6.41 &4.15 &0.0 &41.48\\
Advent~\cite{vu2019advent} & AT &87.10 &42.27 &80.57 &18.46 &12.90 &21.69 &24.88 &82.22 &31.87 &79.82 &65.88 &74.53 &6.83 &7.32 &0.0 &42.42 \\
\hline
CBST~\cite{zou2018domain} & ST &27.09 &27.00 &80.83 &13.88 &2.55 &24.01 &18.10 &82.95 &48.69 &82.84 &61.29 &73.29 &23.71 &5.91 &0.0 &38.14 \\
CRST~\cite{zou2019confidence} & ST &25.52 &23.71 &81.15 &18.85 &3.02 &25.57 &20.39 &83.01 &46.66 &80.91 &63.06  &74.65 &22.76 &12.38 &0.52 &38.81\\
CCM~\cite{li2020content} & ST & 87.02 & 46.76 & 80.78 & 32.50 & 11.25 & 33.60 & 26.84 & 84.09 & 42.65 & 82.14 & 68.36 & 72.89 & 10.35 & 3.69 & 0.0 & 45.52 \\
FDA~\cite{yang2020fda} & ST & 81.63 & 40.89 &82.38 &50.95 &11.23 &22.94 &28.23 &82.50 &49.43 &81.16 &59.79 &70.73 &11.12 &18.97 &0.0 &46.13\\
\hline
IntraDA~\cite{pan2020unsupervised} & A+S & 89.82 & 49.00 & 83.20 & 26.46 & 15.69 & 21.32 & 23.80 & 83.34 & 42.13 & 74.49 & 68.45 & 75.62 & 8.53 & 11.31 & 0.0 &44.88 \\
CBST*~\cite{zou2018domain} & A+S &88.70 &56.66 &82.88 &32.16 &17.59 &32.30 &19.38 &83.80 &40.74 &74.57 &69.68 &69.00 &17.77 &12.32 &00.83 &46.56 \\
CRST*~\cite{zou2019confidence} & A+S &88.73 &56.64 &83.04 &31.66 &17.79 &33.89 &19.02 &83.98 &40.00 &75.00 &70.82 &69.60 &18.48 &12.34 &1.38 &46.83\\
IAST~\cite{mei2020instance} & A+S &\textbf{91.35} &\textbf{62.29} &85.55 &35.75 &\textbf{19.62} &36.51 &27.26 &86.10 &\textbf{38.98} &84.36 &\textbf{70.89} &73.03 &13.33 &14.43 &0.90 &49.36 \\
\rowcolor{lightgray} Ours & A+S &91.23 &53.99 &\textbf{85.86} &\textbf{35.90} &18.73 &\textbf{43.43} &\textbf{34.97} &\textbf{86.28} &36.09 &\textbf{86.01} &65.61 &\textbf{81.81} &\textbf{21.74} &\textbf{34.91} &\textbf{2.04} &\textbf{51.91} \\
\hline
Oracle & - &94.98 &67.68 &87.42 &61.10 &32.20 &40.71 &56.11 &86.82 &55.76 &87.47 &70.22 &89.55 &51.65 &73.43 &0.71 &63.72 \\
\hline
\hline
\end{tabular}}
\vspace{-1mm}
\caption{\textbf{Image semantic segmentation results(mIoU).}}
\vspace{-2mm}
\label{tab:domain_adoption_mIoU}
\end{table*}

\begin{table*}[t]
\centering
\resizebox{\textwidth}{!}{\begin{tabular}{l | c | c c c c c c c c c c c c c c c | c }
\hline 
\hline
\multicolumn{18}{c}{VIPER $\rightarrow$  Cityscapes-VPS [$VPQ^{s}$]} \\
\hline
Method &Arch. &RD &SW & BD & FN & PL &TL & TS &Veg. &TR  &SKY & PR & Car& Truck &Bus &Motor & mIOU(\%) \\
\hline
Source-only & - &28.55 &3.84 &65.95 &0.45 &0.0 &0.1 &5.37 &75.22 &5.50 &64.29 &18.38 &48.50 &1.34 &0.0 &0.0 &21.17  \\
\hline
AdaptSegNet~\cite{tsai2018learning} & AT &84.97 &18.90 &75.87  &4.66  &0.05  &2.98 &8.02  &74.76 &6.69 &65.26 &20.85 &59.02 &2.19 &0.0 &0.0 &28.28\\
Advent~\cite{vu2019advent} & AT &85.55 &20.02 &74.59 &6.90 &0.27 &3.31 &8.04 &74.70 &6.66 &65.64 &22.80 &65.84 &2.03 &0.14 &0.0 &29.10 \\
\hline
CBST~\cite{zou2018domain} & ST &5.56 &8.60 &74.88 &7.05 &0.05 &3.14 &5.27 &75.53 &12.31 &67.83 &21.71 &61.30 &4.72 &0.0 &0.0 &23.20 \\
CRST~\cite{zou2019confidence} & ST &5.63 &6.51 &76.27 &5.89 &0.07 &3.21 &5.74 &75.29 &11.06 &64.11 &24.73 &62.37 &6.42 &1.56 &0.0 &23.26 \\
CCM~\cite{li2020content} & ST & 85.59 & 28.19 & 75.71 & 8.34 & 0.10 & 3.44 & 8.88 & 77.60 & 8.81 & 70.81 & 27.54 & 62.54 & 5.06 & 0.0 & 0.0 & 30.84 \\
FDA~\cite{yang2020fda} & ST & 78.78 & 22.03 &77.92 &16.90 & 0.35 & 0.44 & 5.70 & 75.76 & 11.96 & 67.15 & 25.29 & 58.08 & 3.77 & 1.45 & 0.0 & 29.70 \\
\hline
IntraDA~\cite{pan2020unsupervised} & A+S & 89.44 & 29.12 & 78.26 & 9.02 & 0.97 & 0.79 & 7.62 & 78.18 & 11.19 & 62.35 & 28.26 & 65.56 & 5.76 & 0.26 & 0.0 & 31.12 \\
CBST*~\cite{zou2018domain} & A+S &87.67 &41.12 &78.66 &10.39 &0.88 &6.47 &3.86 &77.97 &10.28 &51.52 &29.53 &56.88 &5.36 &0.0 &0.0 &30.71 \\
CRST*~\cite{zou2019confidence} & A+S &87.74 &41.80 &78.81 &9.90 &1.03 &8.16 &3.63 &78.18 &10.87 &52.07 &30.02 &57.50 &5.58 &0.0 &\textbf{0.35} &31.04\\
IAST~\cite{mei2020instance} & A+S &90.06 &\textbf{45.86} &82.54 &\textbf{14.63} &\textbf{2.60} &6.98 &8.00 &80.78 &\textbf{14.84} &70.21 &\textbf{34.56} &62.84 &5.51 &1.04 &0.19 &34.71\\
\rowcolor{lightgray} Ours & A+S &\textbf{90.64} &37.32 &\textbf{82.68} &11.31 &2.46 &\textbf{10.76} &\textbf{14.40} &\textbf{81.20} &13.97 &\textbf{75.42} &32.68 &\textbf{74.89} &\textbf{9.30} &\textbf{9.37} &0.0 &\textbf{36.43} \\
\hline
Oracle & - &94.81 &49.98 &84.24 &23.16 &4.37 &7.10 &28.58 &80.40 &17.73 &73.84 &32.84 &78.63 &10.57 &35.00 &0.0 &41.42 \\
\hline
\hline
\end{tabular}}
\vspace{-1mm}
\caption{\textbf{Video semantic segmentation results($VPQ^{s}$).}}
\vspace{-4mm}
\label{tab:domain_adoption_VPQ}
\end{table*}

\subsubsection{Clip Adaptive Pseudo Label Generation}
As consecutive frames share a large amount of redundant information, we define each clip after \textit{Aggregate} process as a single instance. The threshold of current clip $\theta_{t}^{(k)}$, for class k, is defined as follows:
\begin{equation}
\begin{split}
\theta_{t}^{(k)} = \beta\theta_{t-1}^{(k)} + (1-\beta)\Psi(x_{T}, \theta_{t-1}^{(k)}), \\
\Psi(x_{T}, \theta_{t-1}^{(k)}) = \widetilde{\mathbbm{U}}_{T}^{(k)}[\alpha{\theta_{t-1}^{(k)}}^{\gamma}|\widetilde{\mathbbm{U}}_{T}^{(k)}|]
\end{split}
\label{eqn:CAST_select}
\end{equation}
The same threshold $\theta_{t}^{(k)}$ is adopted to each frame, which is exponential moving average between threshold of previous clip $\theta_{t-1}^{(k)}$ and $\Psi(x_{T}, \theta_{t-1}^{(k)})$. $\Psi(x_{T}, \theta_{t-1}^{(k)})$ is acquired similar to IAST, while the operating unit is sequence of predictions rather than a prediction of single image. The overall pseudo label generation is summarized in \algref{alg:pseudo}.

\noindent\textbf{Full objective function for VST} Overall loss function is expressed as
\begin{equation}
\begin{split}
L_{VST}(G)= \sum_{l=t,t-\tau}[L_{seg}(P_{T,l},RF(\hat{y}_{T,l})) \\
+ \lambda_{reg}L_{reg}(P_{T,l},RF(\hat{y}_{T,l}))]
\end{split}
\label{eqn:total_VST}
\end{equation}

\subsubsection{Pseudo Label Refinement via Temporal Consensus}
Corresponding pixels across the video should belongs to the same class. However, in practice, it is not always held on the pseudo labels. To eliminate such noise in pseudo label level, we implement pseudo label refinement with temporal correspondence. Specifically, reference pseudo label $\hat{y}_{T,t-\tau}$ is warped onto current pseudo label $\hat{y}_{T,t}$ and check the consensus between them. If not, the model is not learned from the region. The refined current pseudo label $RF(\hat{y}_{T,t})$ can be formulated as follow:

\begin{equation}
RF(\hat{y}_{T,t}) = \mathbbm{1}_{[\hat{y}_{T,t} = W_{t-\tau \Rightarrow t}(\hat{y}_{T,t-\tau})]}\hat{y}_{T,t}
\label{eqn:refine}
\end{equation}

\section{Experiments}
\subsection{Experimental settings}
\subsubsection{Datasets}
We evaluate our video domain adaptation framework on the synthetic-to-real scenarios: VIPER~\cite{richter2017playing} to Cityscapes-VPS~\cite{kim2020vps}. The VIPER dataset has 254,064 video frames and corresponding semantic labels. We only take 
42455 images on ``day" conditions as training data. Cityscape-VPS is video-level extension of Cityscapes dataset~\cite{Cordts2016Cityscapes}, which further annotate 5 frames out of each 30-frames video clip. Following official split in ~\cite{kim2020vps}, we use 400 clips to train a model and evaluated on 50 validation sequences. 

\begin{figure*}[t]
    \centering 
    \includegraphics[width=0.95\textwidth]{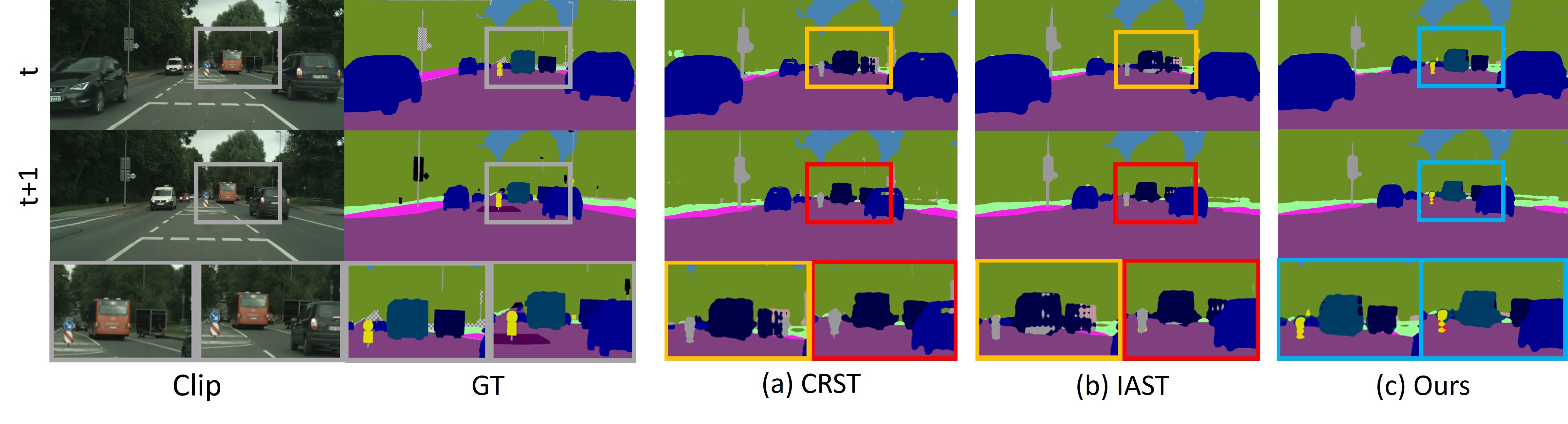}
    \vspace{-3mm}
    \caption{\textbf{Qualitative comparison with the baselines.} We indicate yellow and red boxes for inaccurate and inconsistent prediction, respectively. We can see that previous approaches suffer from both wrong and temporally inconsistent prediction. Instead, our framework successfully resolves both issues. Best viewed in color.
    }
    \label{fig:qual_result}
    \vspace{-3mm}
\end{figure*}

\subsubsection{Baselines}
We compare our methods with several baselines. (1) Source-only is a model naively trained on source domain. (2) Adversarial based image UDA methods include AdaptSegNet~\cite{tsai2018learning}, Advent~\cite{vu2019advent} which rely solely on GAN based adversarial training. (3) Self-training based image UDA methods, CBST~\cite{zou2018domain}, CRST~\cite{zou2019confidence}, CCM~\cite{li2020content} and FDA~\cite{yang2020fda}, iteratively generate pseudo-labels and retrain the models. (4) Advanced methods such as IntraDA~\cite{pan2020unsupervised} and IAST~\cite{mei2020instance} combine adversarial and self-training. In addition, we build strong baselines, CBST* and CRST*, by applying each methods on top of adversarially adapted model that is used in IAST.

\subsubsection{Evaluation Metric}
The predicted results should be both accurate as groundtruth and consistent in time. We first evaluate image-level semantic segmentation performance using standard mean intersection of union(mIoU). In addition, we borrow the video panoptic quality(VPQ)~\cite{kim2020vps} to measure the video-level accuracy. To fit the task of video semantic segmentation, we view the prediction of each class as single instance (i.e. stuff) when the video panoptic quality is calculated. We call the modified VPQ metric $VPQ^{s}$ in this paper. We measure mIoU and $VPQ^{s}$ on 15 common classes.

\subsubsection{Implementation details}
Motivated by ~\cite{liu2020efficient}, we adopt image semantic segmentation models and process each video frame independently for inference.
We use a pre-trained ResNet-50~\cite{he2015deep} as backbone network and adapt DeepLabv2~\cite{chen2017deeplab} as semantic segmentation head. Due to the limitation of GPU memory, we randomly select 2 consecutive frames during training. 
From the image discriminator architecture in ~\cite{tsai2018learning,vu2019advent}, we replace the first 2D convolution layer with 3D convolution layer of temporal stride 2. This modification allows the discriminator $D_{V}$ to take sequence of predictions as input. We set $th$ to 0.7 for cutting out the occluded area.

\subsection{Comparison with State-of-the-art}
In ~\tabref{tab:domain_adoption_mIoU} and ~\tabref{tab:domain_adoption_VPQ}, we report the quantitative adaptation performance on VIPER to Cityscapes-VPS. We compare proposed method with Source-only model, state-of-the-art image UDA-models including adversarial based~\cite{tsai2018learning, vu2019advent}, self-training based~\cite{zou2018domain, zou2019confidence}, and combined methods~\cite{mei2020instance}.

For image-level evaluation, proposed method shows best mIoU 51.91\% with healthy margin. For example, we boost the state-of-the-art image-level UDA method, IAST, by +2.55\% mIoU. It implies that per-frame semantic segmentation quality is improved by effectively leveraging spatio-temporal context in training time. Furthermore, as shown in the video metric and \figref{fig:qual_result}, we can clearly observe that our model produce both accurate and consistent prediction compared to the baselines. Note that proposed method introduce no computational overhead in the inference time.

\subsection{Ablation Study on VAT}
We show the contribution of proposed component to VAT performance in \tabref{tab:Abl_VAT}. 

\begin{table}[t]
\centering
\resizebox{0.4\textwidth}{!}
{
\def\arraystretch{1.1}
\begin{tabular}{c|cccc|cc}
\hline
\multicolumn{1}{c}{} & \multicolumn{2}{c}{Dice} & \multicolumn{2}{c}{Adv.} & \multicolumn{2}{c}{Metric}  \\
Method & Single & Tube & Img. & Vid. & mIoU & $VPQ^{s}$ \\
\hline
\hline
(1)  &            &             &            &                 & 34.02  & 21.17 \\
(2)  &            &             & \checkmark &                 & 43.95  & 29.78 \\
(3)  & \checkmark &             & \checkmark &                 & 44.03  & 29.51 \\
(4)  &            & \checkmark  & \checkmark &                 & 44.75  & 30.27 \\
Ours &            & \checkmark  &            &  \checkmark     & \textbf{45.53}  & \textbf{30.81} \\
\hline
\end{tabular}
}
\caption{\textbf{Ablation study on video adversarial training.} We empirically verify the effectiveness of proposed tube matching loss and sequence discriminator.}
\label{tab:Abl_VAT}
\end{table}
\begin{table}[t]
\centering
\resizebox{0.48\textwidth}{!}
{
\def\arraystretch{1.1}
\begin{tabular}{lcccccc}
\hline
Method & ST & Agg. & Reg. & Ref. & mIoU & $VPQ^{s}$ \\
\hline
\hline
Video Adversarial Training  &    &       &   &       & 45.53  & 30.81 \\
\hline
Class Balanced~\cite{zou2018domain} & \checkmark  &       &    &      & 47.63  & 32.26 \\
Instance Adaptive~\cite{mei2020instance} & \checkmark  &       &    &      & 47.98  & 32.31 \\
\hline
Clip Adaptive & \checkmark  &             &            &                 & 49.18  & 33.48 \\
+Aggregated prediction  & \checkmark &  \checkmark  &  &        & 49.52  & 34.02 \\
+Regularization  &  \checkmark   & \checkmark  & \checkmark &   & 51.52  & 35.78 \\
+Temporal Refinement &  \checkmark   & \checkmark  &  \checkmark  &  \checkmark     & \textbf{51.90}  & \textbf{36.43} \\
\hline
\end{tabular}
}
\caption{\textbf{Ablation study on video self-training.} ``Agg.", ``Reg." and ``Ref." denote temporally aggregated prediction, regularization and online pseudo label refinement, respectively.}
\vspace{-3mm}
\label{tab:Abl_VST}
\end{table}

\subsubsection{Effectiveness of Tube Matching Loss.} 
We first investigate the impact of the tube matching loss.
In model-(4), We add tube matching loss on top of the image-level adaptation model(model-(2)). 
As improved image and video metrics indicate, we confirm that tube-matching loss helps to learn spatio-temporal knowledge in source domain, which results in powerful video domain knowledge transfer to target domain.
For more comprehensive insight, we further report the result of model-(3) that involve image-level degradation of tube matching loss (e.g. $L_{tube}(P_{S,i},y_{S,i})$).
In terms of $VPQ^{S}$, model-(3) presents even worse performance than the simple image-level adaptation model.
It implies that the improvements are truly based on the tube-level design of loss, not just from using the image-level dice coefficient.

\begin{figure*}[t]
    \centering 
    \includegraphics[width=0.98\textwidth]{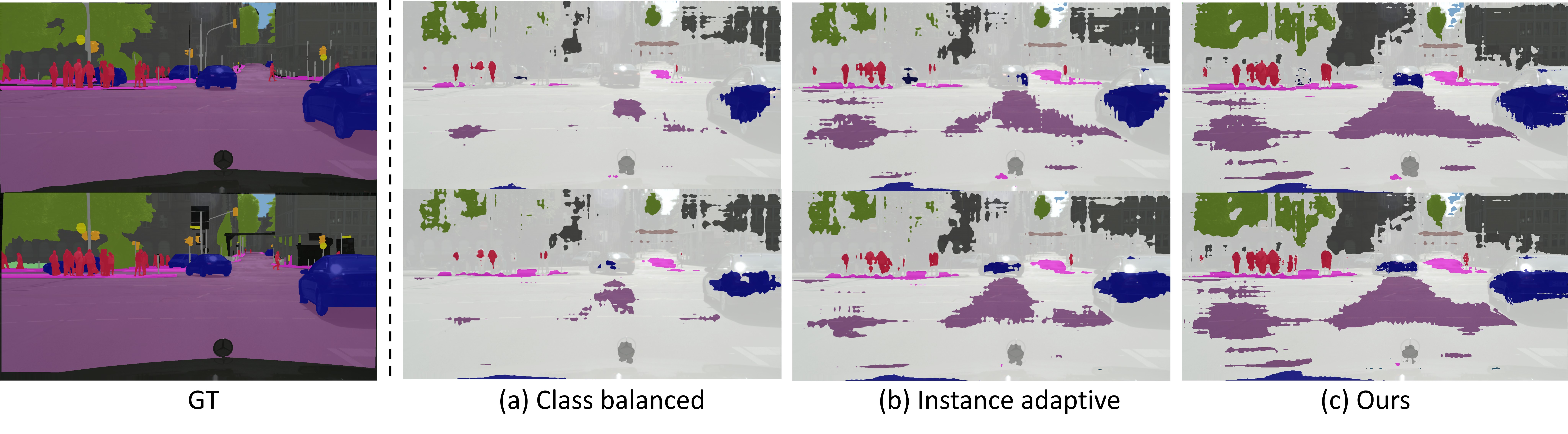}
    \caption{\textbf{Visual comparisons on pseudo labels.} We can clearly observe that proposed method generate more accurate and consistent pseudo labels over the baselines.}
    \label{fig:pseudo label}
    \vspace{-3mm}
\end{figure*}

\subsubsection{Effectiveness of Sequence Discriminator.}
We also explore the efficacy of the sequence discriminator. 
Our final VAT model in \tabref{tab:Abl_VAT}, further adopts sequence discriminator instead of image discriminator. 
We empirically confirm that the sequence discriminator allows better adaptation by additionally leveraging temporal information. 
As a result, our VAT model achieves 1.5 mIoU and 1.0 $VPQ^{s}$ over the prior adversarial method, which provides better pseudo labels to the self-training phase.

\subsection{Ablation Study on VST}
\subsubsection{Effectiveness of Proposed Pseudo Label Generation Strategy.}
Here, we study the efficacy of clip adaptive pseudo label generation.
We compare it with the previous state-of-the-art image-based pseudo label generation approaches: class-balanced~\cite{zou2018domain}, instance adaptive~\cite{mei2020instance}.
As shown in the \tabref{tab:Abl_VST}, the model trained with the proposed clip adaptive pseudo labels shows the best performance.
Plus, adopting the presented `temporally aggregated prediction' further improves the performance. Finally, we also provide qualitative results in the \figref{fig:pseudo label}. For example,  both ``people'' and ``car'' classes are overlooked in previous approaches but densified with our method.
Quantitatively and qualitatively, we clear see that our clip adaptive pseudo label generation generates more accurate and temporally consistent labels, which ultimately increase the performance of self training.

\subsubsection{Importance of video adversarial training.}
\begin{figure}[t]
    \centering 
    \includegraphics[width=0.45\textwidth]{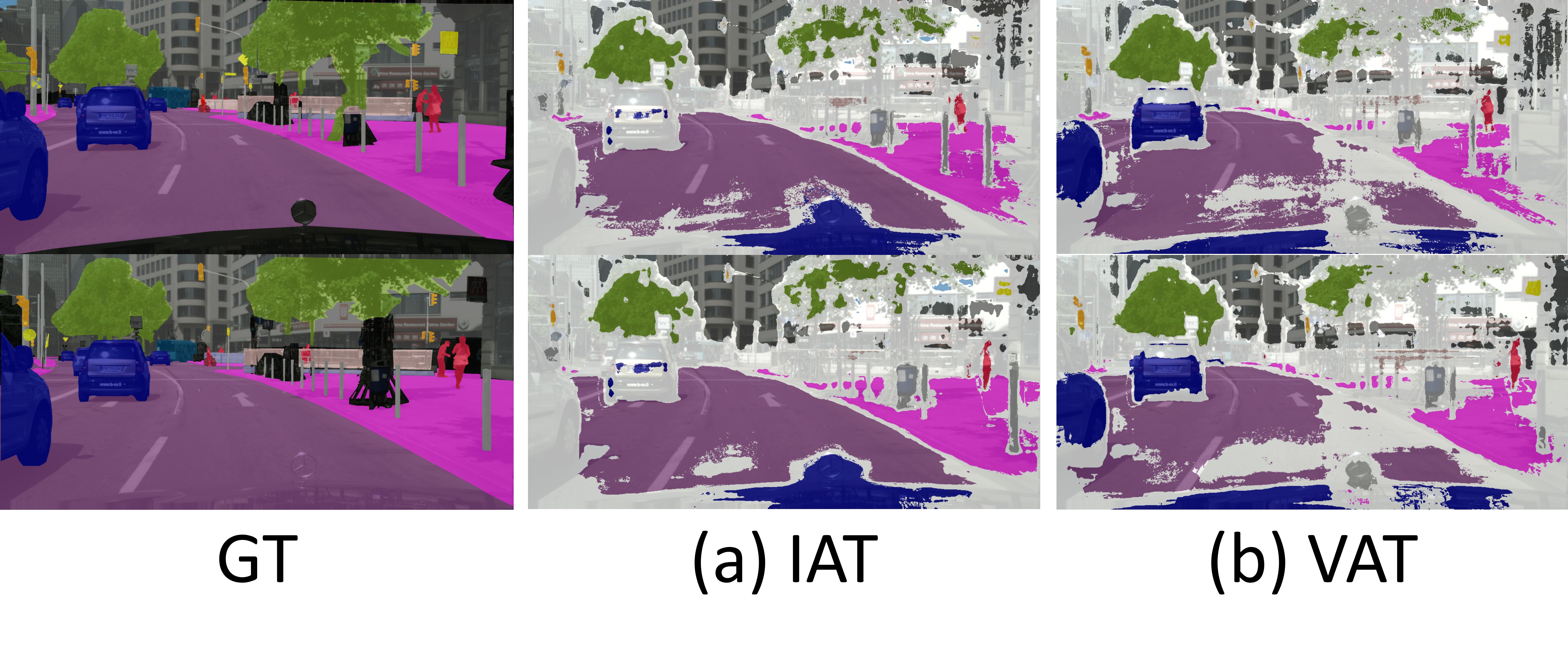}
    \caption{\textbf{Pseudo labels generated from different adversarial models.} It is obvious that the quality of generated pseudo labels depend on the pre-trained adversarial model since ours with VAT model shows better visual effect on pseudo labels than ours with IAT model.}
    \label{fig:adv_ablat}
    \vspace{-3mm}
\end{figure}

Here we study how different first-phase adversarial methods affect the final performance.
We compared them using the same video self-training in the second-phase.
The results are summarized in~\tabref{tab:Abl_warmup}.
Without any proposals in VAT, the final performance drop by a significant margin, showing the efficacy of our proposals. 
The best performance is achived when all the proposals (i.e., tube matching and video discriminator) are used together.
This again shows the importance of exploring the video context in designing the adaptation technique.
We also provide qualitative pseudo labels when using IAT and VAT in~\figref{fig:adv_ablat}.
We obtain much dense and accurate pseudo labels with the video-level adversarial pre-training.

\subsubsection{Visual analysis of online pseudo label refinement.}
We illustrate the role of online pseudo label refinement in the \figref{fig:online_qual}. It removes noisy pseudo labels by checking temporal consensus. For example, original pseudo labels mislabel bicycle regions as a person, while the refinement process successfully eliminates such parts. As a result, this process prevents the model from constantly fitting to noise, showing the better performance (\tabref{tab:Abl_VST}).

\begin{table}[t]
\centering
\resizebox{0.4\textwidth}{!}
{
\def\arraystretch{1.1}
\begin{tabular}{cc|cc}
\hline
Adversarial Training & Self-Training & mIoU & $VPQ^{s}$ \\
\hline
\hline

(2)  &  Ours                &  48.58   &  33.95   \\
(3)  &  Ours                &  50.18   &  34.83   \\
(4)  &  Ours                &  50.70   &  35.66   \\
Ours &  Ours                &  \textbf{51.90}   &  \textbf{36.43}   \\
\hline
\end{tabular}
}
\caption{\textbf{Importance of video adversarial training.} We run our full VST phase on the different adversarial models.}
\label{tab:Abl_warmup}
\end{table}

\section{Conclusion}

In this paper, we explore a new domain adaptation task, unsupervised domain adaptation for video semantic segmentation. 
We present a novel framework that consists of two video-specific domain adaptive training: VAT and VST.
In the first step, we distill source knowledge using standard cross entropy loss and the newly presented tube matching loss jointly. 
Meanwhile, the VAT is applied for the feature alignment.
In the second step, the video self-training is used for target data learning. 
The temporal information is leveraged to generate dense, accurate pseudo labels.
We significantly outperform all the previous strong image-based UDA baselines on Viper to CityscapeVPS scenario.
We hope many follow-up studies being presented in the future based on our proposals and results.

\newpage

\appendix
\addcontentsline{toc}{section}{Appendices}

\section*{Appendices}

In this supplementary section, we provide,
\begin{enumerate}[label=\Alph*.]
    \item Additional evaluations on other datasets,
    \item Comparisons with ~\cite{chen2020generative}
    \item Details of a new UDA framework that supports multiple popular UDA approaches in a unified platform,
    \item The impact of online pseudo label refinement,
    \item Additional analyses of the pseudo labels,
    \item Limitations and discussions of the proposals,
    \item More qualitative adaptation results on Cityscape-VPS.
\end{enumerate}

\section{A. Additional Evaluations on Other Datasets}
To show that our proposal is indeed general across the datasets, we additionally evaluate our method and baselines on two more benchmarks: (a) weather adaptation: CamVid-Sunny to CamVid-Foggy and (b) synthetic-to-real: SYNTHIA-Seq to Cityscape-VPS. 

For CamVid~\cite{brostow2009semantic}, it contains 5 video sequences of a total of 701 annotated frames. we use 4 sunny video as the source domain and 1 foggy video as the target domain. We compare the proposed method with the state-of-the-art baselines in the ~\tabref{tab:domain_adoption_mIoU_camvid} and ~\tabref{tab:domain_adoption_VPQ_camvid}.
Synthia-Seq~\cite{ros2016synthia} has 7 video sequences under different scenarios and traffic conditions. Among them, We only take 1928 images captured on virtual city as source domain. Experimental results are shown in the ~\tabref{tab:domain_adoption_mIoU_syn} and ~\tabref{tab:domain_adoption_VPQ_syn}.
We can see that our proposal consistently achieves the state of the art on the proposed benchmarks.

\section{B. Comparisons with ~\cite{chen2020generative}}
In the main paper, we claim that their framework shows inferior performance on complex video semantic segmentation.
To provide experimental evidence, we conduct quantitative comparisons with ~\cite{chen2020generative} on CamVid-Sunny to CamVid-Foggy.
For fair comparisons, we run our method in the similar setting with ~\cite{chen2020generative}. Specifically, we adopt PSPNet~\cite{zhao2017pyramid} as a segmentation model wit ResNet-50 backbone. The results are shown in the~\tabref{tab:comp_chen}. The proposed method outperforms ~\cite{chen2020generative} with healthy margins.

\section{C. New UDA framework}
\label{section:framework}
One of the paper's main contributions is presenting a new UDA framework that supports multiple popular and contemporary UDA approaches for semantic segmentation.
Specifically, our framework contains eight strong UDA baselines in a unified platform. A list is given as follows,
\subsection{Adversarial-based}
\begin{itemize}
    \item AdaptSegNet~\cite{tsai2018learning}: adapting structured output semantic segmentation logits, proposed in 2018.
    \item Advent~\cite{vu2019advent}: adapting entropy maps, proposed in 2019. 
\end{itemize}
\subsection{Self-training-based}
\begin{itemize}
    \item CBST~\cite{zou2018domain}: class-balanced pseudo label generation, proposed in 2018.
    \item CRST~\cite{zou2019confidence}: confidence regularized self-training, proposed in 2019. 
    \item FDA~\cite{yang2020fda}: fourier domain adaptation for semantic segmentation, proposed in 2020.
    \item CCM~\cite{li2020content}: content-consistent matching for domain adaptive semantic segmentation, proposed in 2020.
\end{itemize}
\subsection{Combining above two methods}
\begin{itemize}
    \item IntraDA~\cite{pan2020unsupervised}: unsupervised intra-domain adaptation for semantic segmentation through self-supervision, proposed in 2020. 
    \item IAST~\cite{mei2020instance}: instance adaptive pseudo label generation, proposed in 2020.
\end{itemize}

We believe this framework is by far the first complete UDA toolbox.
Upon this framework, we developed our new proposals, video adversarial training (VAT) and video self-training (VST), and enabled fair apple-to-apple comparisons.
The codes and models will be released.

\section{D. The impact of online pseudo label refinement}
\label{section:refinement}

In the main paper, we exploit the temporal information in pseudo label generation so that the temporally inconsistent labels are \textit{cut-out}.
Another reasonable baseline is to directly borrow the reference information to \textit{fill-in} the missing labels in current frame.
In practice, we propagate the pseudo label information in the reference frame ($\hat{y}_{T,t-\tau}$) to the target frame ($\hat{y}_{T,t}$) using flow-based warping function ($ W_{t-\tau \Rightarrow t}$). The overall procedure can be formulated as,
\begin{equation}
    \begin{split}
    Prop(\hat{y}_{T,t})
    = \mathbb{R}_{T,t} * \hat{y}_{T,t} \\
    + (1-\mathbb{R}_{T,t}) * \mathbb{I}_{t-\tau \Rightarrow t} * W_{t-\tau \Rightarrow t}(\hat{y}_{T,t-\tau})
    \end{split}
\label{eqn:temp_prop}
\end{equation}
where $\mathbb{R}_{T,t} = \{\boldsymbol{1}|\hat{y}_{t}^{(h,w)} \neq 0\}$ is pseudo label region of current image, $\mathbb{I}_{t-\tau \Rightarrow t}$ is the non-occluded map from $x_{T,t-\tau}$ to $x_{T,t}$ and * represents element-wise multiplication. 

\begin{figure*}[t]
    \centering 
    \includegraphics[width=0.85\textwidth]{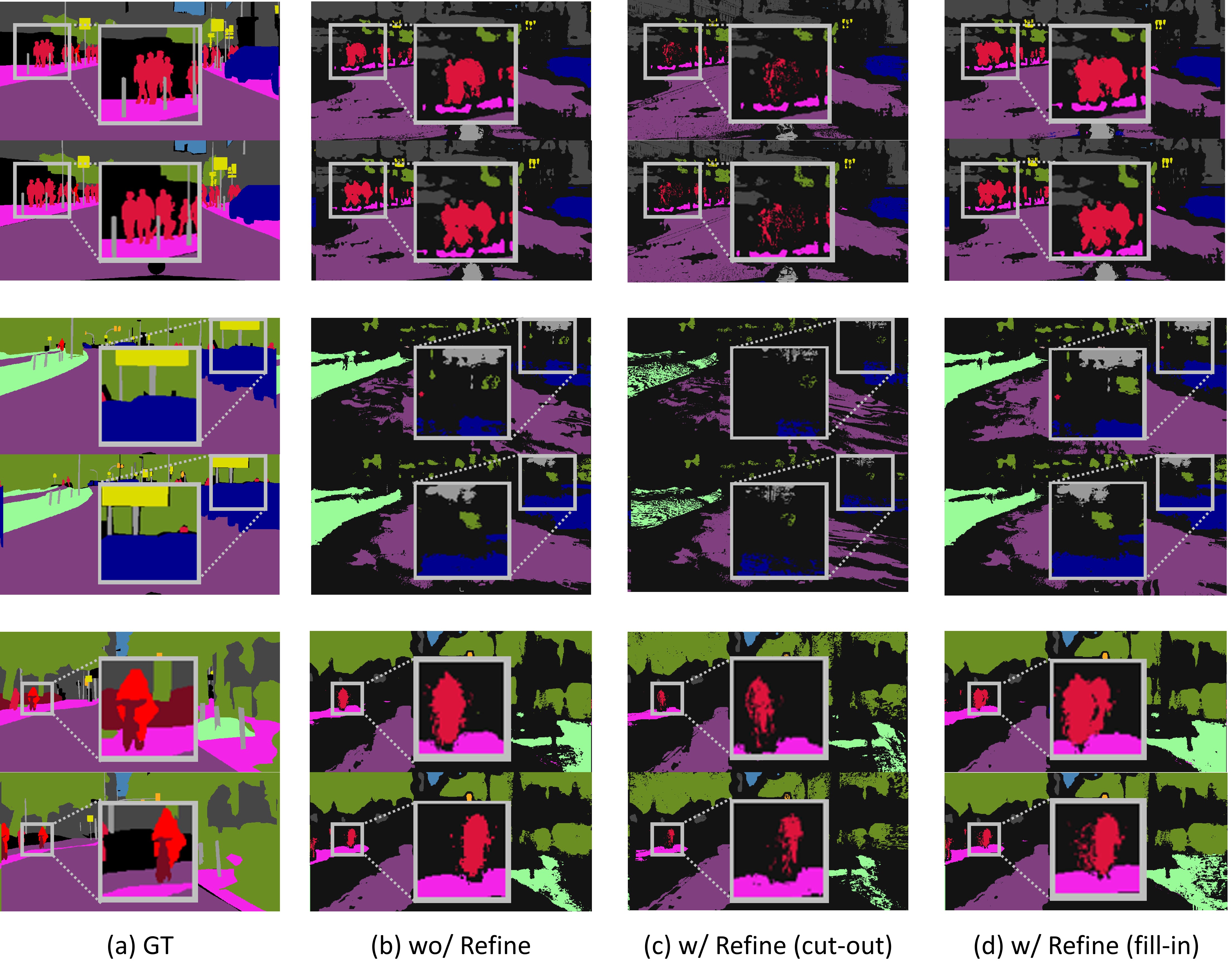}
    \caption{\textbf{Visualization of pseudo-labels w/o and w/ the online refinements.} The proposed \textit{cut-out} refinement successfully eliminate noise on the pseudo labels by checking temporal consensus. However, \textit{fill-in} based method makes additional noise on pseudo labels.}
    \label{fig:online_qual_supple}
\end{figure*}

The comparison results are summarized in~\tabref{tab:Abl_online_refine}.
We observe that \textit{fill-in} approach is inferior to the baseline, whereas the proposed \textit{cut-out} algorithm shows improvement over it.
We see this phenomenon comes from the in-nature noise in the pseudo label, and thus cut-out-based regularization is better than fill-in-based label accumulation.
To back our claim, we also provide qualitative results in~\figref{fig:online_qual_supple}.`
We can observe the general tendency that our cut-out based label refinement less accumulates erroneous labels than the fill-in based approach.

\begin{table}[h]
\centering
\resizebox{0.45\textwidth}{!}
{
\setlength{\tabcolsep}{20pt}
\def\arraystretch{0.95}
\begin{tabular}{lcc}
\hline
Method & mIoU & $VPQ^{s}$ \\
\hline
\hline
Base                & 51.52  & 35.78 \\
Base+fill-in            & 50.0  & 34.78\\
Base+cut-out (ours)     & \textbf{51.91}  & \textbf{36.43}\\
\hline
\hline
\end{tabular}
}
\caption{\textbf{Comparative performance on online refinement.} 
We experiment different online refinement methods on top of proposed VAT and VST.} 
\label{tab:Abl_online_refine}
\end{table}

\section{E. Additional analysis of Pseudo Labels}
In the main paper, we already highlight the effectiveness of the proposed pseudo label generation strategy by showing quantitative and qualitative comparisons in Table 4 and Figure 5. Besides, we measure mIoU of different pseudo labels and name it as P-mIoU. For fair comparisons, We tune the hyperparameter $\alpha$ (see Preliminary in the main paper) of each method to have the same proportion of pseudo labels. The results are in the ~\tabref{tab:P-miou}. We again observe that our proposal generates the most accurate pseudo labels regardless of the proportion. It implies that leveraging the additional temporal information is essential.

\section{F. Qualitative results with Video Demo}
\label{section:demo}
 We release video results~\footnote{\url{https://youtu.be/z-rBcY87XCw}} on test sets of cityscapes dataset~\cite{Cordts2016Cityscapes}. Our method shows much clear and consistent prediction compared to state-of-the-art image UDA method, confirming its robustness and effectiveness.

\section{G. Limitations and Discussions}
\label{section:failure_case}

In order to make the proposed video UDA framework more competitive and facilitate the future research, we point out a few specific items that call for continuing efforts:

\begin{table}[t]
\centering
\resizebox{0.45\textwidth}{!}
{
\setlength{\tabcolsep}{15pt}
\def\arraystretch{0.90}
\begin{tabular}{l|c|ccc}
\hline
\multirow{2}{*}{} & \multirow{2}{*}{Proportion} & \multicolumn{3}{c}{Methods} \\
                  &     & \textbf{CB} & \textbf{IA}  &  Ours   \\
\hline
\multirow{2}{*}{P-mIoU} & 0.3 &  67.8   & 67.9  & \textbf{68.8} \\
                        & 0.4 &  64.8   & 66.4  & \textbf{67.1} \\
\hline
\hline
\end{tabular}
}
\vspace{2mm}
\caption{\textbf{P-mIoU of different pseudo labels.} \textbf{CB} and \textbf{IA} denote class balanced~\cite{zou2018domain} and instance adaptive~\cite{mei2020instance} pseudo label generation methods, respectively.}
\label{tab:P-miou}
\end{table}

\begin{figure*}[t]
    \centering 
    \includegraphics[width=0.85\textwidth]{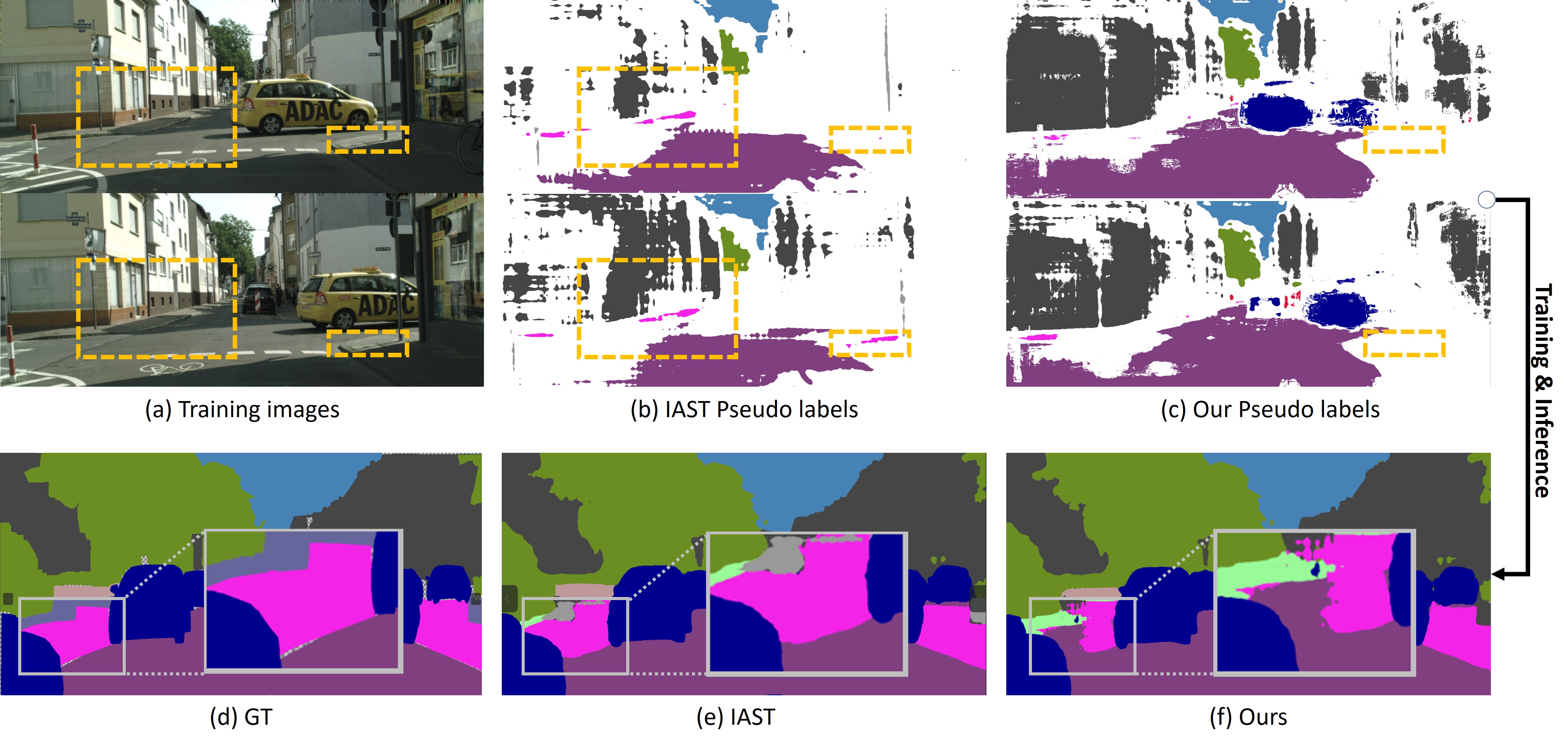}
    \caption{\textbf{Visualization of cause and effect in our failure case.} Our model is comparably weak to certain class, which could be originated from pseudo label generation process. Detail analysis is on Sec.F}
    \label{fig:failure_case}
\end{figure*}

\begin{itemize}
    \item \textbf{FlowNet dependency.} Currently, the overall pseudo label generation in our framework highly relies on the FlowNet~\cite{ilg2016flownet}, which is trained on the external simulated data.
    This inevitably brings inferior adaptation results on certain classes that the FlowNet cannot capture well.
    For example, in~\figref{fig:failure_case}, we see that the adaptation results of ``Sidewalk'' class is inferior to IAST framework~\cite{mei2020instance} (see \figref{fig:failure_case} (d)-(f)), which originates from the imperfect pseudo labels (see \figref{fig:failure_case} (a)-(c)). One possible strategy might be learn to adapt without the FlowNet. We expect deeper explorations in this directions in the future.


    \item \textbf{How to select the initial model for VST.} As pointed out in Figure 6 of the main paper, the quality of the pseudo labels largely depends on the initial model. Even with the same adversarial methods, it is hard to select the best iteration without target domain labels due to the unstable training of adversarial methods.
    Thus, for the better final performance, it is unclear that what criteria to choose the model by. While this issue is an inherent limitation of ``Adversarial-then-Self training" framework itself, it is also important for better video DA methods.
\end{itemize}


\bibliography{egbib}



\begin{table*}
\centering
\resizebox{\textwidth}{!}{\begin{tabular}{l | c | c c c c c c c c c c c | c }
\hline 
\hline
\multicolumn{14}{c}{CamVid-Sunny $\rightarrow$  CamVid-Foggy [mIoU]} \\
\hline
Method &Arch. & Bicyclist & Building & Car & Pole & Fence  & Pedestrian & Road & Sidewalk & Sign & Sky & Tree & mIOU(\%) \\
\hline
Source-only & - & 56.08 & 64.80 & 82.97 & 18.56 & 23.87 & 53.05 & 87.93 & 63.08 & 8.45 & 50.16 & 68.92 & 52.54 \\
\hline
AdaptSegNet~\cite{tsai2018learning} & AT & 47.44 & 78.98 & 74.48 & 12.30 & 26.25 & 43.52 & 82.02 & 54.73 & 4.66 & 86.84 & 77.73 & 53.54\\
Advent~\cite{vu2019advent} & AT & 46.62 & 79.32 & 74.70 & 11.94 & 25.16 & 43.30 & 81.83 & 54.82 & 4.40 & 87.30 & 77.79 & 53.38 \\
\hline
CBST*~\cite{zou2018domain} & A+S & 65.40 & 84.61 & 85.50 & 24.59 & 40.13 & 61.34 & 89.73 & 74.41 & 5.42 & 91.27 & 80.33 & 63.88 \\
CRST*~\cite{zou2019confidence} & A+S & 65.93 & 84.48 & 85.64 & 24.84 & 39.90 & 61.72 & 89.66 & 74.26 & 5.33 & 91.22 & 80.50 & 63.95 \\
IAST~\cite{mei2020instance} & A+S & 65.35 & 85.00 & 86.60 & 21.65 & 39.87 & 60.88 & 90.34 & 72.21 & 7.10 & 91.36 & 80.89 &63.75  \\
\rowcolor{lightgray} Ours & A+S & 69.78 & 85.19 & 86.48 & 24.35 & 40.21 & 61.72 & 91.36 & 76.64 & 10.75 & 91.54 & 81.79 & 65.44 \\
\hline
\hline
\end{tabular}}
\caption{\textbf{Image semantic segmentation results(mIoU) on CamVid-Sunny $\rightarrow$  CamVid-Foggy.}}
\label{tab:domain_adoption_mIoU_camvid}
\end{table*}

\begin{table*}
\centering
\resizebox{\textwidth}{!}{\begin{tabular}{l | c | c c c c c c c c c c c | c }
\hline 
\hline
\multicolumn{14}{c}{CamVid-Sunny $\rightarrow$  CamVid-Foggy [$VPQ^{s}$]} \\
\hline
Method &Arch. & Bicyclist & Building & Car & Pole & Fence  & Pedestrian & Road & Sidewalk & Sign & Sky & Tree & $VPQ^{s}$(\%) \\
\hline
Source-only & - & 28.36 & 48.31 & 79.93 & 0.71 & 3.20 & 26.10 & 87.15 & 51.57 & 3.16 & 32.26 & 50.10 & 37.35 \\
\hline
AdaptSegNet~\cite{tsai2018learning} & AT & 13.62 & 66.26 & 68.68 & 0.0 & 7.92 & 19.55 & 80.53 & 34.95 & 0.0 & 86.74 & 60.67 & 39.90\\
Advent~\cite{vu2019advent} & AT & 12.82 & 66.60 & 68.71 & 0.0 & 7.59 & 19.33 & 80.33 & 34.93 & 0.0 & 87.23 & 60.73  & 39.84 \\
\hline
CBST*~\cite{zou2018domain} & A+S & 38.05 & 74.75 & 83.02 & 2.01 & 23.85 & 37.11 & 88.97 & 71.22 & 1.81 & 91.26 & 70.32 &52.94 \\
CRST*~\cite{zou2019confidence} & A+S & 38.00 & 74.93 & 83.17 & 2.01 & 23.25 & 37.98 & 88.85 & 71.17 & 1.81 & 91.20 & 70.87 &53.02 \\
IAST~\cite{mei2020instance} & A+S & 38.55 & 76.44 & 83.82 & 1.66 & 22.48 & 38.16 & 89.69 & 68.57 & 3.72 & 91.41 & 71.32 & 53.26 \\
\rowcolor{lightgray} Ours & A+S & 46.93 & 76.86 & 83.47 & 2.07 & 23.79 & 39.40 & 90.72 & 73.29 & 5.20 & 91.63 & 72.75 & 55.10 \\
\hline
\hline
\end{tabular}}
\caption{\textbf{Video semantic segmentation results($VPQ^{s}$) on CamVid-Sunny $\rightarrow$  CamVid-Foggy.}}
\label{tab:domain_adoption_VPQ_camvid}
\end{table*}

\begin{table*}[t]
\centering
\resizebox{\textwidth}{!}{\begin{tabular}{l | c | c c c c c c c c c c c | c }
\hline 
\hline
\multicolumn{14}{c}{SYNTHIA-Seq $\rightarrow$  Cityscapes-VPS [mIoU]} \\
\hline
Method &Arch. & road & sidewalk & building & pole & light & sign & vegetation & sky & person & rider & car & mIOU(\%) \\
\hline
Source-only & - & 79.00 & 34.31 & 72.30 & 23.51 & 11.36 & 28.13 & 72.08 & 72.75 & 57.82 & 10.46 & 64.24 & 47.81 \\
\hline
AdaptSegNet~\cite{tsai2018learning} & AT & 85.29 & 40.14 & 77.26 & 22.08 & 9.03 & 27.96 & 72.12 & 68.06 & 58.93 & 0.0 & 75.49 & 48.76 \\
Advent~\cite{vu2019advent} & AT & 84.02 & 37.50 & 76.54 & 21.52 & 8.22 & 27.56 & 71.76 & 68.17 & 57.40 & 0.0 & 75.24 &47.99  \\
\hline
CBST*~\cite{zou2018domain} & A+S & 87.61 & 43.96 & 77.81 & 32.46 & 23.43 & 35.59 & 75.34 & 79.68 & 56.03 & 10.03 & 81.44 & 54.85  \\
CRST*~\cite{zou2019confidence} & A+S & 86.07 & 43.77 & 79.12 & 31.17 & 27.47 & 38.94 & 77.11 & 80.47 & 59.48 & 12.47 & 82.43 & 56.23 \\
IAST~\cite{mei2020instance} & A+S & 88.34 & 48.68 & 80.42 & 34.19 & 28.88 & 39.51 & 79.74 & 77.27 & 59.30 & 9.36 & 83.05 & 57.16  \\
\rowcolor{lightgray} Ours & A+S & 89.84 & 50.51 & 81.46 & 32.98 & 30.27 & 40.39 & 80.91 & 81.13 & 59.87 & 8.79 & 84.40 & 58.23  \\
\hline
\hline
\end{tabular}}
\caption{\textbf{Image semantic segmentation results(mIoU) on SYNTHIA-Seq $\rightarrow$  Cityscapes-VPS benchmark.}}
\label{tab:domain_adoption_mIoU_syn}
\end{table*}

\begin{table*}[t]
\centering
\resizebox{\textwidth}{!}{\begin{tabular}{l | c | c c c c c c c c c c c | c }
\hline 
\hline
\multicolumn{14}{c}{SYNTHIA-Seq $\rightarrow$  Cityscapes-VPS [$VPQ^{s}$]} \\
\hline
Method &Arch. & road & sidewalk & building & pole & light & sign & vegetation & sky & person & rider & car & $VPQ^{s}$(\%) \\
\hline
Source-only & - & 76.02 & 16.89 & 62.67 & 0.64 & 1.49 & 6.40 & 59.59 & 61.39 & 17.10 & 0.0 & 50.62 & 32.07 \\
\hline
AdaptSegNet~\cite{tsai2018learning} & AT & 84.59 & 19.15 & 66.34 & 0.10 & 0.25 & 6.14 & 57.95 & 59.21 & 17.95 & 0.0 & 62.90 & 34.05 \\
Advent~\cite{vu2019advent} & AT & 83.23 & 15.89 & 65.44 & 0.04 & 0.17 & 5.68 & 56.80 & 59.43 & 17.14 & 0.0 & 62.40 & 33.29 \\
\hline
CBST*~\cite{zou2018domain} & A+S & 84.98 & 28.85 & 70.48 & 2.29 & 1.79 & 9.78 & 64.58 & 67.74 & 24.17 & 0.0 & 69.47 &38.56  \\
CRST*~\cite{zou2019confidence} & A+S & 83.49 & 33.70 & 71.79 & 1.01 & 2.11 & 13.56 & 69.23 & 69.08 & 23.33 & 0.0 &73.11 & 40.04 \\
IAST~\cite{mei2020instance} & A+S & 85.74 & 37.25 & 73.36 & 4.42 & 3.07 & 13.89 & 73.08 & 70.12 & 26.79 & 0.0 & 73.34 & 41.92  \\
\rowcolor{lightgray} Ours & A+S & 88.15 & 39.20 & 75.04 & 4.17 & 4.36 & 14.89 & 74.93 & 73.10 & 28.75 & 0.0 & 73.70 & 43.30  \\
\hline
\hline
\end{tabular}}
\caption{\textbf{Video semantic segmentation results($VPQ^{s}$) on SYNTHIA-Seq $\rightarrow$  Cityscapes-VPS benchmark.}}
\label{tab:domain_adoption_VPQ_syn}
\end{table*}


\begin{table*}
\centering
\resizebox{\textwidth}{!}{\begin{tabular}{l | c | c c c c c c c c c c c | c }
\hline 
\hline
\multicolumn{14}{c}{CamVid-Sunny $\rightarrow$  CamVid-Foggy [mIoU]} \\
\hline
Method &Arch. & Bicyclist & Building & Car & Pole & Fence  & Pedestrian & Road & Sidewalk & Sign & Sky & Tree & mIOU(\%) \\
\hline
Source-only & - & 4.19 & 35.34 & 35.55 & 8.31 & 2.14 & 22.63 & 54.07 & 45.70 & 9.63 & 5.17 & 37.61 & 23.49 \\
VideoGAN~\cite{chen2020generative} & - & 8.98 & 50.44 & 37.24 & 15.27 & 2.84 & 31.62 & 58.33 & 62.63 & 3.45 & 49.68 & 62.77 & 34.18 \\
\hline
Source-only$\dagger$ & - & 5.91 & 50.37 & 33.42 & 0.03 & 0.0 & 0.1 & 44.86 & 22.48 & 0.02 & 47.15 & 37.90 & 22.02 \\
\rowcolor{lightgray} Ours & A+S & 16.76 & 70.81 & 44.94 & 0.49 & 19.5 & 17.22 & 41.23 & 41.07 & 0.0 & 86.31 & 69.01 & 43.82 \\
\hline
\hline
\end{tabular}}
\caption{\textbf{Quantitative comparisons of ours with ~\cite{chen2020generative} on CamVid-Sunny $\rightarrow$  CamVid-Foggy.} $\dagger$ denotes the source only model reproduced in this paper.}
\label{tab:comp_chen}
\end{table*}


\end{document}